\documentclass{article}

\usepackage{microtype}
\usepackage{graphicx}
\usepackage{booktabs} 

\usepackage[utf8]{inputenc} 
\usepackage[T1]{fontenc}    
\usepackage{url}            
\usepackage{booktabs}       
\usepackage{amsfonts}       
\usepackage{nicefrac}       
\usepackage{microtype}      
\usepackage{xcolor}         
\usepackage{amsmath}

\DeclareMathOperator*{\argmin}{arg\,min}
\usepackage{amssymb} 
\usepackage{epstopdf}
\usepackage{array}
\usepackage{multirow}
\usepackage{bm}
\usepackage{tikz}
\usepackage{enumitem}
\usepackage{caption}
\usepackage{subcaption}
\usepackage[subtle]{savetrees}

\usepackage{hyperref}

\usepackage[accepted]{icml2021}

\icmltitlerunning{Generative network-based reduced-order model for prediction, data assimilation and uncertainty quantification}

\begin{document}

\twocolumn[
\icmltitle{Generative network-based reduced-order model for prediction, data assimilation and uncertainty quantification}



\icmlsetsymbol{equal}{*}

\begin{icmlauthorlist}
\icmlauthor{Vinicius L. S. Silva}{amcg}
\icmlauthor{Claire E. Heaney}{amcg}
\icmlauthor{Nenko Nenov}{}
\icmlauthor{Christopher C. Pain}{amcg}
\end{icmlauthorlist}

\icmlaffiliation{amcg}{Applied Modelling and Computation Group, Imperial College London, UK}

\icmlcorrespondingauthor{Vinicius L. S. Silva}{v.santos-silva19@imperial.ac.uk}

\icmlkeywords{Machine Learning, ICML}

\vskip 0.3in
]



\printAffiliationsAndNotice{}  

\begin{abstract}
We propose a new method in which a generative network (GN) integrate into a reduced-order model (ROM) framework is used to solve inverse problems for partial differential equations (PDE). The aim is to match available measurements and estimate the corresponding uncertainties associated with the states and parameters of a numerical physical simulation. 
The GN is trained using only unconditional simulations of the discretized PDE model. 
We compare the proposed method with the golden standard Markov chain Monte Carlo. 
We apply the proposed approaches to a spatio-temporal compartmental model in epidemiology. The results show that the proposed GN-based ROM can efficiently quantify uncertainty and accurately match the measurements and the golden standard, using only a few unconditional simulations of the full-order numerical PDE model. 
\end{abstract}

\section{Introduction}\label{sec:intro}
Complex physics and engineering systems are usually described in terms of partial differential equations (PDE) that, for most problems of practical interest, cannot be solved analytically. Then it is necessary to use numerical methods to solve the governing equations. The discretization of these equations is usually performed using finite difference, finite volume, finite element or a combination of these procedures \citep{golub:92,ames:14}. 
Nonetheless, these methods need a large number of degrees of freedom to solve the PDEs accurately. This fact can generate prohibitively expensive simulations in terms of computational time and memory demand. Furthermore, these models are built on limited information, which makes their predictions uncertain. The actual values of the model states and parameters are not known, and usually measurements are sparse in space and/or time. Therefore, it is necessary to assimilate observed data (calibrate model states and parameters in order to generate results that match the measurements) and formally propagate the uncertainties through the forward numerical simulator. 
In this context, reduced-order models (ROM) are computationally appealing and have been attracting significant attention in the last decades \citep{Rozza:08,cardoso:09,hesthaven2018non,xiao:19}. The aim of a ROM is to reduce the computational burden of numerical simulations by creating a low-dimensional representation of a high-dimensional model or discretized system.

In order to quantify the uncertainty in the states and parameters of a numerical PDE simulation, one requires multiple random models that match or are conditional to measurements. After simulating the conditional models, an empirical distribution of the variables of interest can be obtained \citep{liu:03,lakshminarayanan2017simple}. The validity of the uncertainty quantification depends on the quality of the generated conditional simulations. Nonetheless, it is often difficult and computationally expensive to generate a single conditional model (that honours the measurements), suggesting that the task of quantifying uncertainty must be even more difficult \citep{liu:03,sudret:17,cacuci2019berru}. It is unfeasible to use methods such as rejection sampling (RS) and Markov chain Monte Carlo (MCMC) to propagate uncertainty through most practical PDE simulations due to their computational cost \citep{oliver:97,liu:03,oliver:11b,stordal:18}. Therefore, approximate methods need to be used. Among them,  \citet{liu:03} showed that the randomized maximum likelihood (RML) \citep{kitanidis:95,oliver:96}, also called randomize-then-optimize (RTO) \citep{bardsley:14}, performed better than other approximate methods. 

In this paper, we propose a new method inspired by the RML (or RTO) in which a generative network (GN) within a ROM framework is used to quantify the uncertainty of a numerical PDE simulation in the presence of measurements. The ROM uses a low-dimensional space for the spatial distribution of the simulation states. Then the GN is used to learn the evolution of the low-dimensional states over time. 
The GN is trained using only unconditional simulations of the full-order numerical model. After training, the GN-based ROM can be used to predict the evolution of the spatial distribution of the simulation states and observed data can be assimilated. Here, we describe the process required in order to quantify uncertainty, during which 
no additional simulations of the full-order numerical model are required. We apply these methods to estimate the uncertainty in the states and parameters of a spatio-temporal compartmental model in epidemiology, that was constructed in order to represent the spread of COVID-19 in an idealized town. 



The source code, data, hardware configuration, used in this work are available at \url{https://github.com/viluiz/gan}.

\section{Test case description}\label{sec:testcase}

The test case used here is the spatio-temporal variation of a virus infection in an idealized town. The extended SEIRS model used in this work \citep{silva:21a,quilodran:21} extends the traditional theory of the dynamics of infectious diseases \cite{anderson:92,bjornstad:18,bjornstad:20b} to account for variations not only in time but also in space. The PDEs describing the test case can be found in Appendix~\ref{sec:appseirs}.

\subsection{Extended SEIRS model and problem set up}
\label{sec:eseirs}

\begin{figure}[!tb]
	\centering
	\includegraphics[width=1.0\columnwidth]{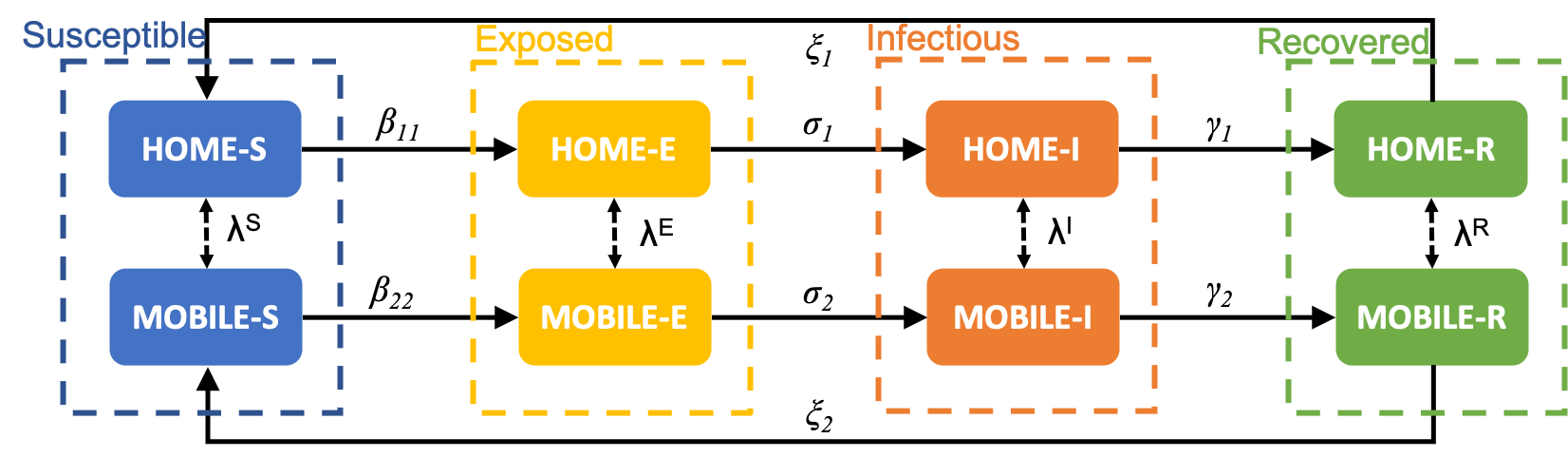}
	\caption{Diagram of the extended SEIRS model. The diagram shows how people move between groups and compartments at a given point in space (or one cell in the grid). The vital dynamics and the transport via diffusion are not displayed here.}
	\label{fig:ExtSEIRSmodel}
\end{figure}

The extended SEIRS model used in this work consists of four compartments (Susceptible - Exposed - Infections - Recovered) and two people groups (Home - Mobile). Figure \ref{fig:ExtSEIRSmodel} shows the diagram of how individuals move between compartments and groups. The model starts with some individuals in the infectious compartments (Home-I/Mobile-I). The members of these compartment will  spread the pathogen to the susceptible compartments (Home-S/Mobile-S). Upon being infected, the members of the susceptible compartments are moved to the exposed compartments (Home-E/Mobile-E) and remain there until they become infectious. Infectious individuals remain in the infectious compartment until they become recovered (Home-R/Mobile-R). Recovered people can also become susceptible again due to the loss of immunity. 

One important factor in dynamics of infectious diseases is the basic reproduction number ($\mathcal{R}_0$), which represents the expected number of new cases caused by a single infectious member in a completely susceptible population \citep{dietz:93,hethcote:00}. The $\mathcal{R}_0$ controls how rapidly the disease could spread. 
Here, we have two $\mathcal{R}_{0}$, one for each group of people. $\mathcal{R}_{0\,1}$ representing the basic reproduction number of people at home, and $\mathcal{R}_{0\,2}$ representing the basic reproduction number of people that are mobile and outside their homes therefore. For this case, we can also calculate an Effective $\mathcal{R}_{0}$ representing the $\mathcal{R}_{0}$ seen by the whole population at a specific time (see Appendix~\ref{sec:appseirs}). 


The idealized town and problem set up used in this work are the same as in \citet{silva:21a} and \citet{quilodran:21}. The simulation comprises 8000 people in an area of $100\times100$km, discretized on a regular grid. The initial condition is that 0.1\% of people have been exposed to the virus and will thus develop an infection. The model parameters are the two $\mathcal{R}_{0\,h}$ (home and mobile), and the model states are the number of people in each compartment (S, E, I and R) for each people group and for each grid block. Further information about the discretization and solution methods of the full-order numerical simulation can be found in \citet{silva:21a} and \citet{quilodran:21}.

\section{Method}\label{sec:method}



\begin{figure}[htb]
	\centering
	\includegraphics[width=1.0\columnwidth]{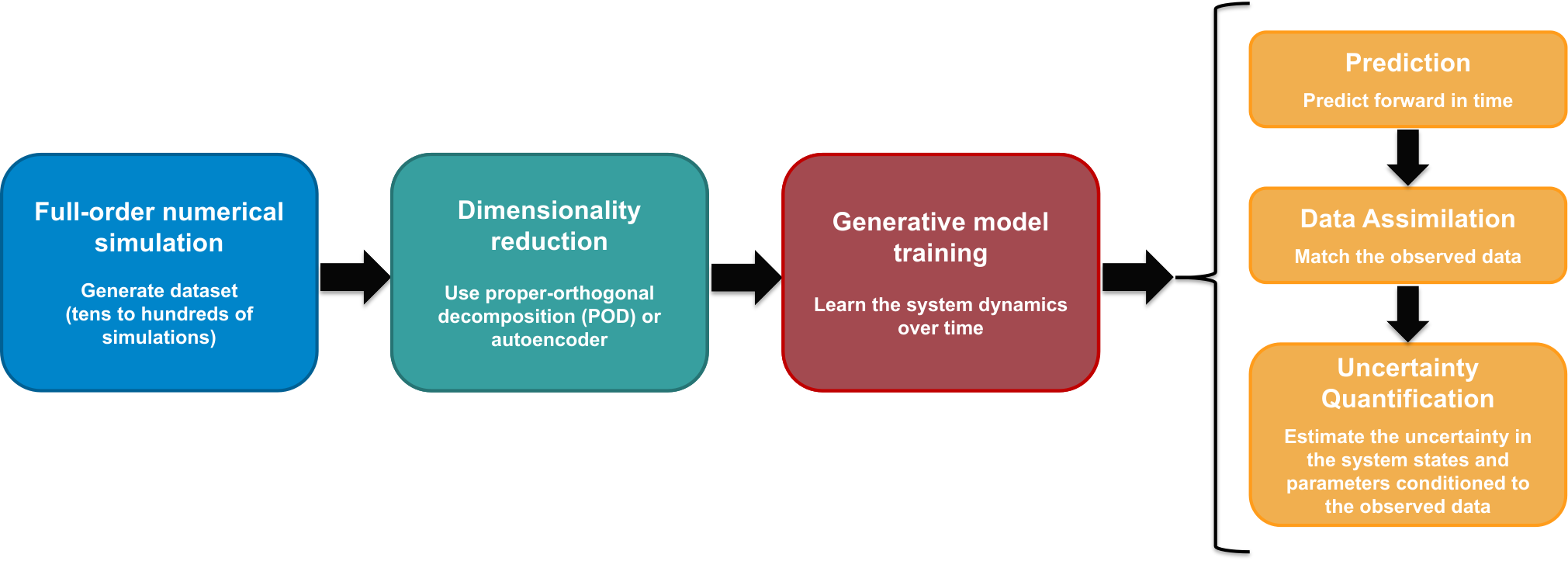}
	\caption{Proposed framework to quantify the uncertainty in the states and parameters of numerical PDE simulations.}
	\label{fig:framework}
\end{figure}

We propose a novel framework to quantify uncertainties of numerical PDE simulations. The aim is to condition the numerical simulation to observed data (usually sparse in space and/or time) and estimate the corresponding uncertainties in the model states and parameters. Figure~\ref{fig:framework} shows the proposed framework. We start by generating the dataset, running the full-order numerical simulations with different model parameters. The number of runs used to generate the dataset is orders of magnitude smaller than what would be required, if we exclusively rely on the full-order simulations to calculate the uncertainties. After generating the dataset, we apply the dimensionality reduction to the model states at each time step. Having the compressed states and corresponding model parameters, we can train a generative model to learn the dynamic of the system over time. The goal is for the network to learn how to generate a sequence (in time) of physically plausible compressed states and the corresponding model parameters. Following the training of the generative network, we can use it to predict forward in time \citep{quilodran:21,silva:21a}, to find one solution that honour available measurements \citep{silva:21a}, and/or to find multiple solutions that honour the measurements and also generates the uncertainties in the model states and parameters (as proposed here). 

\begin{figure}[htb]
	\centering
	\includegraphics[width=1.0\columnwidth]{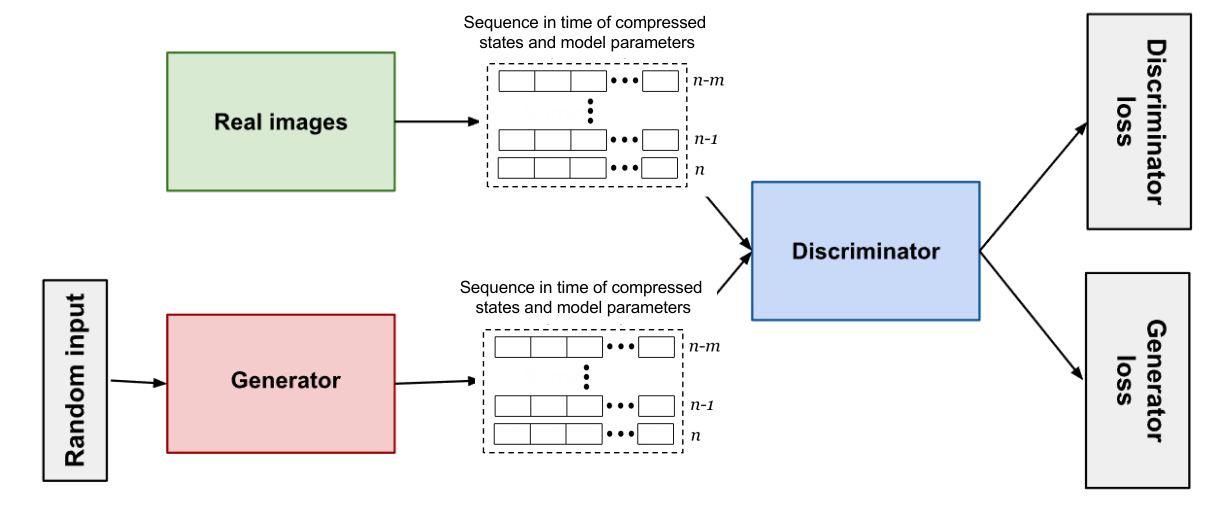}
	\caption{The generative model is trained to generate a sequence in time of compressed states and model parameters from the numerical PDE simulation.}
	\label{fig:gan}
\end{figure}

In this work, we train a generative adversarial network (DCGAN \citep{radford:15} with non-saturating loss \citep{goodfellow2016nips}) to produce time-sequences of the compressed states and model parameters of the extended SEIRS model, from a random distribution as input (see Figure~\ref{fig:gan}). The compression of the model states is performed using proper orthogonal decomposition. After training, the prediction (PredGAN method) is performed in a recurrent way as described in Appendix~\ref{sec:pred}, the data assimilation (DA-PredGAN method) modifies the prediction process to include the observed data (see Appendix~\ref{sec:da}), and the uncertainty quantification (UQ-PredGAN method proposed here) modifies the data assimilation process to work with an ensemble of models instead of a single realization. Further details about the proposed uncertainty quantification method can be found in Appendix~\ref{sec:uq}.   

\section{Results and discussion}\label{sec:results}

In this section, we apply the proposed framework to solve an inverse problem for the extended SEIRS model. The goal is to quantify the uncertainties in the model states and parameters, considering the availability of observed data. The model represents the spread of COVID-19 in an idealized town (as in \citet{quilodran:21} and \citet{silva:21a}). We generate the observed data (``measurements'') from a full-order numerical simulation ($\mathcal{R}_{0\,1}=7.7$, $\mathcal{R}_{0\,2}=17.4$) that was not included in the training set. Observed data was collected at five points of the domain and the measurements are available every two days. We measured only infectious and recovered people, as in \citet{silva:21a}. The $\mathcal{R}_{0\,h}$ are not used as observed data. For generating the priors (unconditional simulations) 200 model parameters  $\mathcal{R}_{0\,h}$ were sampled from a normal distribution with a mean of 10 and standard deviation of 4. The mean and standard deviation were chosen based on \citet{kochanczyk:20}. The 200 model parameters and their corresponding initial conditions were used to start the UQ-PredGAN process. For each of the model parameters, one data assimilation was performed as described in Appendix~\ref{sec:uq}. After the data assimilation, 121 realizations were accepted based on their data mismatch error. It is worth noting that for the whole uncertainty quantification process using the UQ-PredGAN, we required only 40 full-order numerical simulations (for training the GAN). 

\begin{figure}[!tb]
	\centering
	\includegraphics[width=1.0\columnwidth]{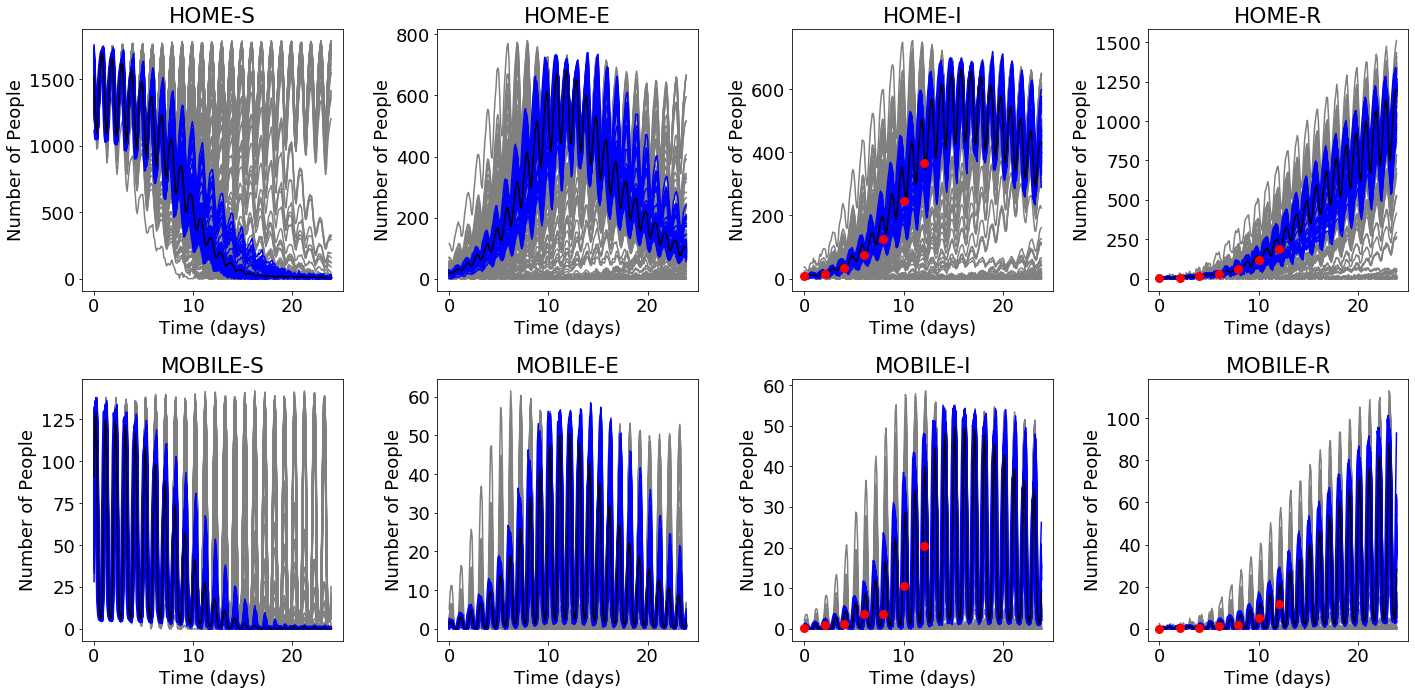}
	\label{fig:prior}
	\caption{Results of the UQ-PredGAN for each group and compartment at one point in space. The red dots represent the observed data (``measurements''), the grey lines the unconditional simulations (priors), the blue lines the conditional simulations (posteriors), and the black line the posterior mean.}
	\label{fig:priorvspost}
\end{figure}

Figure \ref{fig:priorvspost} shows the UQ-PredGAN results for each group and compartment (model states) at one point in space where observed data was available. The priors (grey lines) are the first forward march of each data assimilation, and the posteriors (blue lines) are the last forward march of the accepted realizations. The posterior mean (black line) is also shown in the plots. We can see from these figures that the conditional simulations (posteriors) generated by the UQ-PredGAN match the observed data, within some tolerance (we considered a measurement error of 5\%), and the uncertainty is propagated through the simulation time. The high frequency oscillation presented in the results corresponds to a daily cycle, when mobile people leave their homes during the day and return to them at night. Comparable results were observed at other points in domain, hence they are not presented here.  

\begin{figure}[!tb]
	\centering
	\begin{subfigure}[t]{0.30\columnwidth}
		\centering
		\includegraphics[width=\columnwidth]{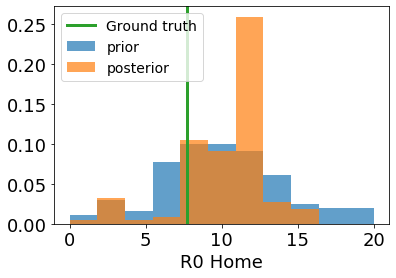}
		\label{fig:prior}
	\end{subfigure}
	\begin{subfigure}[t]{0.30\columnwidth}
		\centering
		\includegraphics[width=\columnwidth]{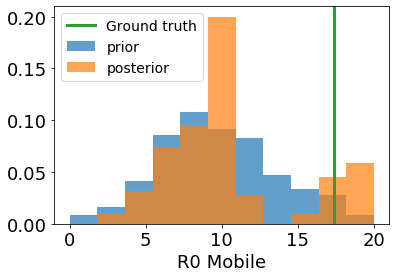}
		\label{fig:post}
	\end{subfigure}
	\begin{subfigure}[t]{0.30\columnwidth}
		\centering
		\includegraphics[width=\columnwidth]{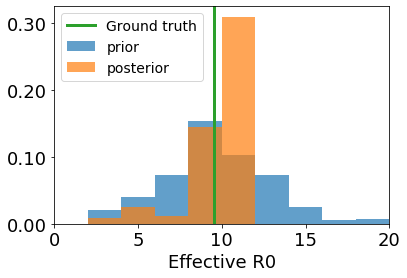}
		\label{fig:post}
	\end{subfigure}   
	\begin{subfigure}[t]{0.30\columnwidth}
		\centering
		\includegraphics[width=\columnwidth]{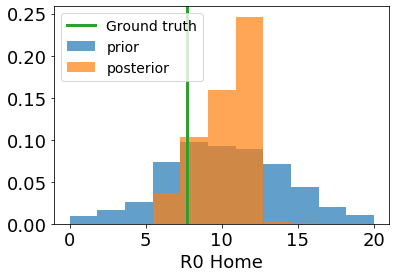}
		\label{fig:prior}
	\end{subfigure}
	\begin{subfigure}[t]{0.30\columnwidth}
		\centering
		\includegraphics[width=\columnwidth]{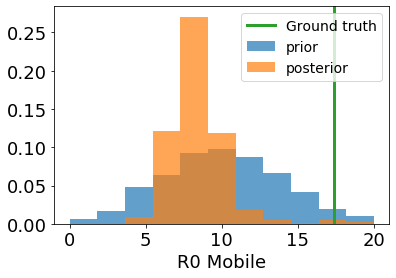}
		\label{fig:post}
	\end{subfigure}
	\begin{subfigure}[t]{0.30\columnwidth}
		\centering
		\includegraphics[width=\columnwidth]{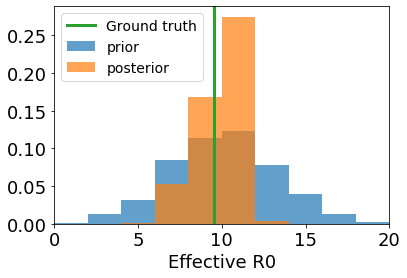}
		\label{fig:post}
	\end{subfigure} 	
	\caption{Probability density functions of the $\mathcal{R}_{0\,h}$ and Effective $\mathcal{R}_0$ at day 12. Comparison of UQ-PredGAN (first row) with the golden standard MCMC (second row). The ground truth represents the values in the full-order numerical simulation used to generate the observed data. }
	\label{fig:pdfcompR0}
\end{figure}

In the plots from the first row of Figure \ref{fig:pdfcompR0}, we show the probability density function (PDF) of the $\mathcal{R}_{0\,h}$ (model parameters) and Effective $\mathcal{R}_{0}$, for the priors and posteriors. The result shows that the UQ-PredGAN was able to reduce the uncertainty in the model parameters approaching the true values used to generate the observed data. Note that the data assimilation is an inverse and usually ill-posed problem, hence different values of $\mathcal{R}_{0\,h}$ could match the measurements within some tolerance. We have also run a comparison of the UQ-PredGAN with the golden standard MCMC (using the Metropolis–Hastings algorithm). The results from the MCMC are in the plots from the second row of Figure \ref{fig:pdfcompR0}. It is worth mentioning that to run the MCMC, it took more than a month on a dedicated workstation (more than 100,000 samples and we stopped because of time limitations, see Appendix~\ref{sec:apphard} for the hardware configuration), while to run the UQ-PredGAN only 12 hours (two days if we add the GAN training). We can notice from Figure \ref{fig:pdfcompR0} that the UQ-PredGAN generates uncertainties that have a reasonable match with the golden standard, but with orders of magnitude less computational time. In practical terms, it is unaffordable to run the MCMC for most numerical PDE simulations.

We can notice that the distributions generated by the UQ-PredGAN have longer ``tails'' than the ones from the MCMC. We checked the results from the simulations in that portion of the distribution and they reasonably match the observed data. It may indicate that the MCMC has not converged yet, which means that we would need many more simulation samples (>100,000) for the MCMC to generate the correct tails, which makes it impractical.  Figure \ref{fig:pdfcompR0} also shows that the UQ-PredGAN is able to sample the second small mode of the mobile group $\mathcal{R}_{0\,2}$. It is where the true value of the $\mathcal{R}_{0\,2}$ used to generate the observed data occurs. We can also see that for the mobile group, the UQ-PredGAN posterior PDF does not exactly match the MCMC results as well as for the home group. This could be because the number of mobile people is one order of magnitude smaller than the number of people at home, which gives the latter more importance during the data assimilation process, and the relative rate of change of the number of people in the mobile group is much greater than in the home group, thus small perturbations in the former can cause huge relative deviations. 


\begin{figure}[!tb]
    \centering
    \begin{subfigure}[t]{0.35\columnwidth}
    	\centering
    	\includegraphics[width=\columnwidth]{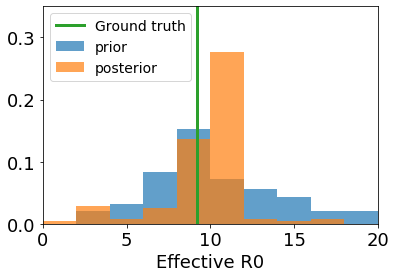}
    	\caption{UQ-PredGAN}
    	\label{fig:post}
    \end{subfigure}
    \begin{subfigure}[t]{0.35\columnwidth}
    	\centering
    	\includegraphics[width=\columnwidth]{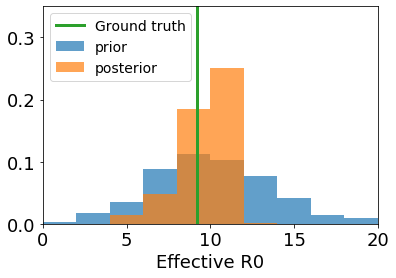}
    	\caption{MCMC}
    	\label{fig:post}
    \end{subfigure}
	\caption{Predicted probability density functions of the Effective $\mathcal{R}_0$ at day 16 (the last observed data was at day 12). Comparison of UQ-PredGAN with the golden standard MCMC. The ground truth represents the value in the high fidelity numerical simulation used to generate the observed data.}
	\label{fig:pdfcompR0time}
\end{figure}

Figure \ref{fig:pdfcompR0time} shows the prediction of the Effective $\mathcal{R}_{0}$ at day 16, along with the ground truth, for the UQ-PredGAN and MCMC. Comparable results were observed for other points in time. The last observed data used in this experiment was at day 12. The results demonstrate that the UQ-PredGAN can generate predictions and uncertainties that accurately match the ground truth and have a reasonable match to the golden standard MCMC. Table \ref{tab:pdfcompR0time} shows the posterior mean, mode and standard deviation (STD) for different days. The mode is calculated based on the histograms. We can notice that going further in time the posterior uncertainty (the STD) in the Effective $\mathcal{R}_0$ increases, in both methods. This is because the further away the predictions are from the last observed data, the greater the uncertainty will be in those predictions. The difference in the standard deviation of the UQ-PredGAN and the MCMC is because it was impractical to run the amount of simulation samples required for the MCMC to represent the tails of the distributions, as already mentioned.    

\begin{table}[!htb]
\caption{Values of the posterior PDF of the Effective $\mathcal{R}_0$ over time. Comparison of UQ-PredGAN with the golden standard MCMC.}
\label{tab:pdfcompR0time}
\vskip 0.15in
\begin{center}
\begin{small}
\begin{sc}
\begin{tabular}{lccc}
\toprule
 Time (days) => & 12 & 16 & 22  \\
\midrule
ground truth     & 9.52 & 9.22 & 9.48 \\
uq-predgan mean   & 9.86 & 9.70 & 9.91 \\
mcmc mean   & 9.89 & 9.68 & 9.92 \\
\midrule
uq-predgan mode   & 11.0 & 11.0 & 11.0  \\
mcmc mode   & 11.0 & 11.0 & 11.0 \\
\midrule
uq-predgan std   & 1.73 & 2.44 & 2.70  \\
mcmc std   & 1.39 & 1.40 & 1.51  \\
\bottomrule
\end{tabular}
\end{sc}
\end{small}
\end{center}
\vskip -0.1in
\end{table}




\section{Conclusion}

We proposed a novel use of generative networks as a reduced-order model, that is able to solve inverse problems for numerical PDE simulations. 
We applied the proposed method to a spatio-temporal compartmental model in epidemiology. The results show that the UQ-PredGAN accurately matches the observed data and efficiently quantifies/reduces uncertainty in the model states (groups and compartments) and model parameters (basic reproduction numbers). We compare the UQ-PredGAN with the golden standard MCMC, and show that it can generate predictions and uncertainties that reasonably match the golden standard, but with orders of magnitude less computational time.   
The proposed method is not limited to the underlying physics of this application, it is a general framework for quantifying uncertainties of numerical physical simulations. 




\bibliography{references}

\begin{thebibliography}{38}
\providecommand{\natexlab}[1]{#1}
\providecommand{\url}[1]{\texttt{#1}}
\expandafter\ifx\csname urlstyle\endcsname\relax
  \providecommand{\doi}[1]{doi: #1}\else
  \providecommand{\doi}{doi: \begingroup \urlstyle{rm}\Url}\fi

\bibitem[Ames(2014)]{ames:14}
Ames, W.~F.
\newblock \emph{Numerical methods for partial differential equations}.
\newblock Academic press, 2014.

\bibitem[Anderson et~al.(1992)Anderson, Anderson, and May]{anderson:92}
Anderson, R.~M., Anderson, B., and May, R.~M.
\newblock \emph{Infectious diseases of humans: dynamics and control}.
\newblock Oxford University Press, 1992.

\bibitem[Bardsley et~al.(2014)Bardsley, Solonen, Haario, and
  Laine]{bardsley:14}
Bardsley, J.~M., Solonen, A., Haario, H., and Laine, M.
\newblock Randomize-then-optimize: A method for sampling from posterior
  distributions in nonlinear inverse problems.
\newblock \emph{SIAM Journal on Scientific Computing}, 36\penalty0
  (4):\penalty0 A1895--A1910, 2014.

\bibitem[Baydin et~al.(2017)Baydin, Pearlmutter, Radul, and Siskind]{baydin:17}
Baydin, A.~G., Pearlmutter, B.~A., Radul, A.~A., and Siskind, J.~M.
\newblock Automatic differentiation in machine learning: a survey.
\newblock \emph{The Journal of Machine Learning Research}, 18\penalty0
  (1):\penalty0 5595--5637, 2017.

\bibitem[Bj{\o}rnstad et~al.(2020)Bj{\o}rnstad, Shea, Krzywinski, and
  Altman]{bjornstad:20b}
Bj{\o}rnstad, O., Shea, K., Krzywinski, M., and Altman, N.
\newblock The {SEIRS} model for infectious disease dynamics.
\newblock \emph{Nature Methods}, 17\penalty0 (6):\penalty0 557--558, 2020.

\bibitem[Bj{\o}rnstad(2018)]{bjornstad:18}
Bj{\o}rnstad, O.~N.
\newblock \emph{Epidemics: Models and data using R}.
\newblock Springer, 2018.

\bibitem[Cacuci(2019)]{cacuci2019berru}
Cacuci, D.~G.
\newblock \emph{BERRU Predictive Modeling: Best Estimate Results with Reduced
  Uncertainties}.
\newblock Springer, 2019.

\bibitem[Cardoso et~al.(2009)Cardoso, Durlofsky, and Sarma]{cardoso:09}
Cardoso, M.~A., Durlofsky, L.~J., and Sarma, P.
\newblock Development and application of reduced-order modeling procedures for
  subsurface flow simulation.
\newblock \emph{International Journal for Numerical Methods in Engineering},
  77\penalty0 (9):\penalty0 1322--1350, 2009.

\bibitem[Chu et~al.(2017)Chu, Zhmoginov, and Sandler]{chu2017cyclegan}
Chu, C., Zhmoginov, A., and Sandler, M.
\newblock {CycleGAN, a Master of Steganography}.
\newblock \emph{arXiv preprint arXiv:1712.02950}, 2017.

\bibitem[Dietz(1993)]{dietz:93}
Dietz, K.
\newblock The estimation of the basic reproduction number for infectious
  diseases.
\newblock \emph{Statistical Methods in Medical Research}, 2\penalty0
  (1):\penalty0 23--41, 1993.

\bibitem[Frid-Adar et~al.(2018)Frid-Adar, Diamant, Klang, Amitai, Goldberger,
  and Greenspan]{frid2018gan}
Frid-Adar, M., Diamant, I., Klang, E., Amitai, M., Goldberger, J., and
  Greenspan, H.
\newblock {GAN-based synthetic medical image augmentation for increased CNN
  performance in liver lesion classification}.
\newblock \emph{Neurocomputing}, 321:\penalty0 321--331, 2018.

\bibitem[Golub et~al.(1992)Golub, Ortega, et~al.]{golub:92}
Golub, G.~H., Ortega, J.~M., et~al.
\newblock \emph{Scientific computing and differential equations: an
  introduction to numerical methods}.
\newblock Academic press, 1992.

\bibitem[Goodfellow(2016)]{goodfellow2016nips}
Goodfellow, I.
\newblock Nips 2016 tutorial: Generative adversarial networks.
\newblock \emph{arXiv preprint arXiv:1701.00160}, 2016.

\bibitem[Goodfellow et~al.(2014)Goodfellow, Pouget-Abadie, Mirza, Xu,
  Warde-Farley, Ozair, Courville, and Bengio]{goodfellow:14}
Goodfellow, I., Pouget-Abadie, J., Mirza, M., Xu, B., Warde-Farley, D., Ozair,
  S., Courville, A., and Bengio, Y.
\newblock Generative adversarial nets.
\newblock In \emph{Advances in neural information processing systems}, pp.\
  2672--2680, 2014.

\bibitem[Goodfellow et~al.(2016)Goodfellow, Bengio, Courville, and
  Bengio]{goodfellow2016deep}
Goodfellow, I., Bengio, Y., Courville, A., and Bengio, Y.
\newblock \emph{Deep learning}, volume~1.
\newblock MIT press Cambridge, 2016.

\bibitem[Hesthaven \& Ubbiali(2018)Hesthaven and Ubbiali]{hesthaven2018non}
Hesthaven, J.~S. and Ubbiali, S.
\newblock {Non-intrusive reduced order modeling of nonlinear problems using
  neural networks}.
\newblock \emph{Journal of Computational Physics}, 363:\penalty0 55--78, 2018.

\bibitem[Hethcote(2000)]{hethcote:00}
Hethcote, H.~W.
\newblock The mathematics of infectious diseases.
\newblock \emph{SIAM review}, 42\penalty0 (4):\penalty0 599--653, 2000.

\bibitem[Karras et~al.(2019)Karras, Laine, and Aila]{karras2019style}
Karras, T., Laine, S., and Aila, T.
\newblock {A Style-Based Generator Architecture for Generative Adversarial
  Networks}.
\newblock In \emph{Proceedings of the IEEE conference on computer vision and
  pattern recognition}, pp.\  4401--4410. 2019.

\bibitem[Karras et~al.(2020)Karras, Laine, Aittala, Hellsten, Lehtinen, and
  Aila]{karras2020analyzing}
Karras, T., Laine, S., Aittala, M., Hellsten, J., Lehtinen, J., and Aila, T.
\newblock {Analyzing and Improving the Image Quality of StyleGAN}.
\newblock In \emph{Proceedings of the IEEE/CVF Conference on Computer Vision
  and Pattern Recognition}, pp.\  8110--8119, 2020.

\bibitem[Kitanidis(1995)]{kitanidis:95}
Kitanidis, P.~K.
\newblock Quasi-linear geostatistical theory for inversing.
\newblock \emph{Water resources research}, 31\penalty0 (10):\penalty0
  2411--2419, 1995.

\bibitem[Kocha{\'n}czyk et~al.(2020)Kocha{\'n}czyk, Grabowski, and
  Lipniacki]{kochanczyk:20}
Kocha{\'n}czyk, M., Grabowski, F., and Lipniacki, T.
\newblock Super-spreading events initiated the exponential growth phase of
  {COVID}-19 with $\mathcal{R}_0$ higher than initially estimated.
\newblock \emph{Royal Society Open Science}, 7\penalty0 (9):\penalty0 200786,
  2020.

\bibitem[Lakshminarayanan et~al.(2017)Lakshminarayanan, Pritzel, and
  Blundell]{lakshminarayanan2017simple}
Lakshminarayanan, B., Pritzel, A., and Blundell, C.
\newblock Simple and scalable predictive uncertainty estimation using deep
  ensembles.
\newblock \emph{Advances in neural information processing systems}, 30, 2017.

\bibitem[Linnainmaa(1976)]{linnainmaa:76}
Linnainmaa, S.
\newblock Taylor expansion of the accumulated rounding error.
\newblock \emph{BIT Numerical Mathematics}, 16\penalty0 (2):\penalty0 146--160,
  1976.

\bibitem[Liu et~al.(2003)Liu, Oliver, et~al.]{liu:03}
Liu, N., Oliver, D.~S., et~al.
\newblock {Evaluation of Monte Carlo methods for assessing uncertainty}.
\newblock \emph{SPE Journal}, 8\penalty0 (02):\penalty0 188--195, 2003.

\bibitem[Liu et~al.(2018)Liu, Qin, Wan, and Luo]{liu2018auto}
Liu, Y., Qin, Z., Wan, T., and Luo, Z.
\newblock Auto-painter: Cartoon image generation from sketch by using
  conditional {W}asserstein generative adversarial networks.
\newblock \emph{Neurocomputing}, 311:\penalty0 78--87, 2018.

\bibitem[Oliver \& Chen(2011)Oliver and Chen]{oliver:11b}
Oliver, D.~S. and Chen, Y.
\newblock Recent progress on reservoir history matching: a review.
\newblock \emph{Computational Geosciences}, 15\penalty0 (1):\penalty0 185--221,
  2011.

\bibitem[Oliver et~al.(1996)Oliver, He, and Reynolds]{oliver:96}
Oliver, D.~S., He, N., and Reynolds, A.~C.
\newblock Conditioning permeability fields to pressure data.
\newblock In \emph{ECMOR V-5th European conference on the mathematics of oil
  recovery}, pp.\  cp--101. European Association of Geoscientists \& Engineers,
  1996.

\bibitem[Oliver et~al.(1997)Oliver, Cunha, and Reynolds]{oliver:97}
Oliver, D.~S., Cunha, L.~B., and Reynolds, A.~C.
\newblock Markov chain monte carlo methods for conditioning a permeability
  field to pressure data.
\newblock \emph{Mathematical geology}, 29\penalty0 (1):\penalty0 61--91, 1997.

\bibitem[Oliver et~al.(2008)Oliver, Reynolds, and Liu]{oliver:08}
Oliver, D.~S., Reynolds, A.~C., and Liu, N.
\newblock \emph{Inverse theory for petroleum reservoir characterization and
  history matching}.
\newblock Cambridge University Press, 2008.

\bibitem[Quilodr{\'a}n-Casas et~al.(2022)Quilodr{\'a}n-Casas, Silva, Arcucci,
  Heaney, Guo, and Pain]{quilodran:21}
Quilodr{\'a}n-Casas, C., Silva, V.~L., Arcucci, R., Heaney, C.~E., Guo, Y., and
  Pain, C.~C.
\newblock {Digital twins based on bidirectional LSTM and GAN for modelling the
  COVID-19 pandemic}.
\newblock \emph{Neurocomputing}, 470:\penalty0 11--28, 2022.

\bibitem[Radford et~al.(2015)Radford, Metz, and Chintala]{radford:15}
Radford, A., Metz, L., and Chintala, S.
\newblock Unsupervised representation learning with deep convolutional
  generative adversarial networks.
\newblock \emph{arXiv preprint arXiv:1511.06434}, 2015.

\bibitem[Rozza et~al.(2008)Rozza, Huynh, and Patera]{Rozza:08}
Rozza, G., Huynh, D. B.~P., and Patera, A.~T.
\newblock Reduced basis approximation and a posteriori error estimation for
  affinely parametrized elliptic coercive partial differential equations.
\newblock \emph{Arch. Comput. Methods Eng}, 15:\penalty0 229--275, 2008.

\bibitem[Silva et~al.(2023)Silva, Heaney, Li, and Pain]{silva:21a}
Silva, V.~L., Heaney, C.~E., Li, Y., and Pain, C.~C.
\newblock {Data Assimilation Predictive GAN (DA-PredGAN) Applied to a
  Spatio-Temporal Compartmental Model in Epidemiology}.
\newblock \emph{Journal of Scientific Computing}, 94:\penalty0 25, 2023.
\newblock \doi{10.1007/s10915-022-02078-1}.

\bibitem[Stordal \& N{\ae}vdal(2018)Stordal and N{\ae}vdal]{stordal:18}
Stordal, A.~S. and N{\ae}vdal, G.
\newblock {A modified randomized maximum likelihood for improved Bayesian
  history matching}.
\newblock \emph{Computational Geosciences}, 22\penalty0 (1):\penalty0 29--41,
  2018.

\bibitem[Sudret et~al.(2017)Sudret, Marelli, and Wiart]{sudret:17}
Sudret, B., Marelli, S., and Wiart, J.
\newblock Surrogate models for uncertainty quantification: An overview.
\newblock In \emph{2017 11th European conference on antennas and propagation
  (EUCAP)}, pp.\  793--797. IEEE, 2017.

\bibitem[Tarantola(2005)]{tarantola:05}
Tarantola, A.
\newblock \emph{Inverse problem theory and methods for model parameter
  estimation}.
\newblock SIAM, 2005.

\bibitem[Wengert(1964)]{wengert:64}
Wengert, R.~E.
\newblock A simple automatic derivative evaluation program.
\newblock \emph{Communications of the ACM}, 7\penalty0 (8):\penalty0 463--464,
  1964.

\bibitem[Xiao et~al.(2019)Xiao, Heaney, Mottet, Fang, Lin, Navon, Guo, Matar,
  Robins, and Pain]{xiao:19}
Xiao, D., Heaney, C.~E., Mottet, L., Fang, F., Lin, W., Navon, I.~M., Guo,
  Y.-K., Matar, O.~K., Robins, A.~G., and Pain, C.~C.
\newblock A reduced order model for turbulent flows in the urban environment
  using machine learning.
\newblock \emph{Building and Environment}, 148:\penalty0 323--337, 2019.

\end{thebibliography}
\bibliographystyle{icml2021}


\clearpage
\appendix

\section{Reduced-order modelling and dimensionality reduction}

The underlying assumption of a ROM is that the solution of the forward model can be accomplished by using considerably fewer degrees of freedom. One of the most widely used model reduction methods is the proper orthogonal decomposition (POD) approach. 
In this work, we use a NIROM with POD as a compression method. We have also tested autoencoders to compress the data; however, the result using POD was practically the same and it does not need training or hyperparameter optimisation. 

After the first stage (the compression), the second stage of the NIROM (the evolution in time of the solutions) is accomplished using a generative network, as shown in Figure \ref{fig:framework}. In this work, we use a generative adversarial network (GAN) as generative model. GANs have received much attention, after achieving excellent results for their generation of realistic-looking images \citep{chu2017cyclegan,frid2018gan,liu2018auto,karras2019style,karras2020analyzing}. Although Long short-term memory (LSTM) networks are widely recognised as one of the most effective sequential models \citep{goodfellow2016deep} for times series predictions, \citet{quilodran:21} compared the performance of the LSTM and GAN as a NIROM. The GAN was able to learn better the underlying data distribution and reduce the forecast divergence, especially when predicting further in time. 

\section{Generative adversarial network training}\label{sec:gantrain}

Proposed by \citet{goodfellow:14}, GANs are unsupervised learning algorithms capable of learning dense representations of the input data and are intended to be used as a generative model. Here, the generator network $G$ directly produces time-sequences of the compressed states and model parameters of a numerical PDE simulation from a random distribution as input (latent vector $\mathbf{z}$).
The discriminator network $D$ attempts to distinguish between samples drawn from the training data and samples drawn from the generator, considered as fake. The output of the discriminator $D(\mathbf{y})$ represents the probability that a sample came from the data rather than a “fake” sample from the generator. Figure \ref{fig:gan} show a schematic of the GAN training. 



In this work, we train the GAN using the non-saturating loss \citep{goodfellow2016nips} and use the DCGAN \citep{radford:15} architecture. 
During the training process the latent space $\mathbf{z}$ is generated as a Gaussian random noise $\mathcal{N}(0,I_L)$, where $I_L$ is an identity matrix of size $L$. The loss function of the discriminator takes as inputs: a time sequence of compressed states and model parameters from the numerical PDE simulation (``real'' sample), and a time sequence of compressed states and model parameters generated by the generator (``fake'' sample). The loss function of the generator only takes the ``fake'' samples as input. After training, the discriminator can be discarded since only the generator is used during the prediction, data assimilation, and uncertainty quantification processes.

\section{PredGAN for time series prediction}\label{sec:pred}

After training the GAN to produce data (compressed simulation states and parameters) at a sequence of $m+1$ time steps, i.e.~given a latent vector $\mathbf{z}$, the output of the generator $G(\mathbf{z})$ will be data at time steps $n-m$ to $n$, no matter which point in time $n$ represents, we can use the generator to make predictions in time in a recurrent way, using a ROM named Predictive GAN (PredGAN) as described in \citet{silva:21a, quilodran:21}. 

\begin{figure}[htb]
	\centering
	\includegraphics[width=1.0\columnwidth]{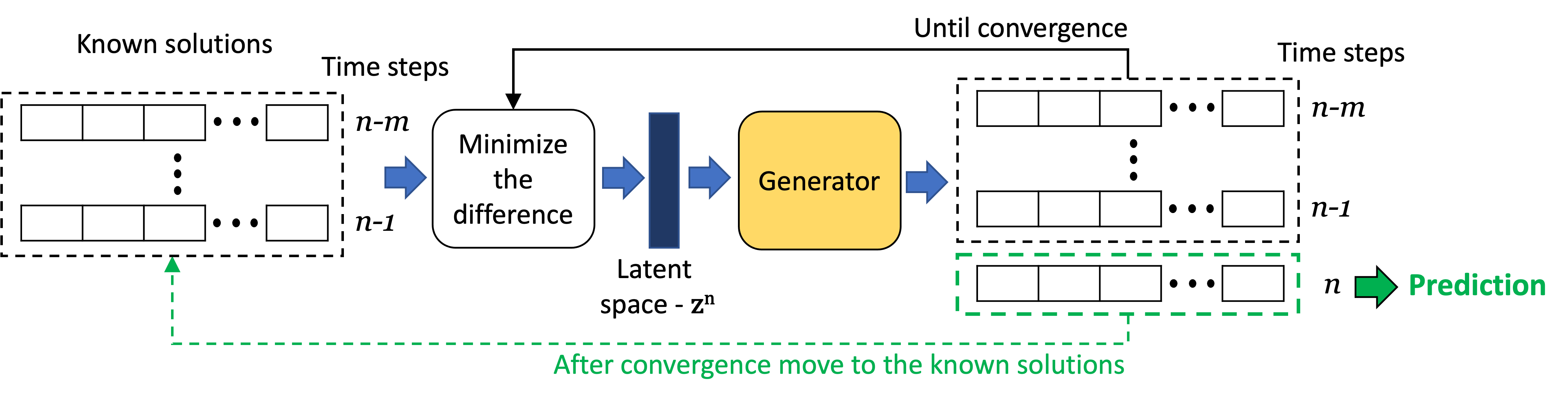}
	\caption{Overview of the PredGAN process.}
	\label{fig:PredGANc}
\end{figure}

Given known solutions at $m$ consecutive time steps, we can perform an optimisation to match the first $m$ time levels in the output of the generator with the known solutions. After convergence, the last time step, $m+1$,  in the output of the generator is the prediction. We now can use this last time level $m+1$ as a known solution and perform another optimisation to predict the time step $m+2$. The process continues until we predict all time steps. Figure \ref{fig:PredGANc} illustrates how the PredGAN works.

In our case, after training, the output of the generator $G(\mathbf{z})$ is made up of $m+1$ consecutive time steps of compressed grid variables~$\bm{\alpha}$ (outputs/states of the numerical PDE simulation), and model parameters~$\bm{\mu}$ (inputs of the numerical PDE simulation). The compressed variables are proper orthogonal decomposition (POD) coefficients, but could also be latent variables from an autoencoder. For a GAN that has been trained with $m+1$ time levels, $G(\mathbf{z})$ takes the following form
\begin{equation}\label{eq:phi}
G(\mathbf{z}^n) = \left[ \begin{array}{c} 
(\bm{\alpha}^{n-m})^T, (\bm{\mu}^{n-m})^T \\[1mm]
\vdots \\[1mm]
(\bm{\alpha}^{n-1})^T, (\bm{\mu}^{n-1})^T \\[1mm]
(\bm{\alpha}^n)^T, (\bm{\mu}^n)^T \\[1mm]
\end{array}
\right]_{(m+1)\ \text{by}\ (N_{\text{POD}}+N_{\mu}) }\,
\end{equation}
where the compressed grid variables are defined as $(\bm{\alpha}^n)^T= [\alpha_1^n, \alpha_2^n, \cdots , \alpha_{N_{\text{POD}}}^n]$. $N_{\text{POD}}$ is the number of principal components, and $\alpha_i^n$ represents the $i$th POD coefficient at time level $n$. The model parameters are represented as $(\bm{\mu}^n)^T =[ \mu_1^n, \mu_2^n, \cdots, \mu_{N_\mu}^n]$. $N_\mu$ is the number of parameters, and $\mu_i^n$ represents the $i$th parameter at time level $n$. 

In each iteration of the PredGAN, one new time step is predicted. Assuming the GAN has already been trained, and we have solutions at time levels from $n-m$ to $n-1$ for the POD coefficients, denoted by $\{\tilde{\bm{\alpha}}^{k}\}_{k=n-m}^{n-1}$, and also have the model parameters over all time steps $\tilde{\bm{\mu}}^{k}$, then to predict the solution at time level $n$ we perform an optimisation defined as
\begin{equation}
\mathbf{z}^{n}  = \argmin_{\mathbf{z^{n}}} \mathcal{L}_p(G(\mathbf{z}^{n})), 
\end{equation}
\begin{equation}\label{eq:loss_prediction}
\begin{gathered}
\mathcal{L}_p(G(\mathbf{z}^{n})) =  \sum_{k=n-m}^{n-1}\, \left(\tilde{\bm{\alpha}}^k - \bm{\alpha}^k \right)^T \bm{W}_\alpha \left(\tilde{\bm{\alpha}}^k - \bm{\alpha}^k \right)   \\
+ \sum_{k=n-m}^{n-1}\,  \zeta_\mu\left(\tilde{\bm{\mu}}^k - \bm{\mu}^k \right)^T \bm{W}_\mu \left(\tilde{\bm{\mu}}^k - \bm{\mu}^k \right),
\end{gathered} 
\end{equation}
where $\bm{W}_\alpha$ is a square matrix of size $N_{POD}$  whose diagonal values are equal to the weights that govern the relative importance of the POD coefficients, all other entries being zero.  $\bm{W}_\mu$ is a square matrix of size $N_{\mu}$ whose diagonal values are equal to the model parameter weights, and the scalar $\zeta_\mu$ controls how much importance is given to the model parameters compared to the compressed variables. The values for all the weighting terms are the same as in \citet{silva:21a}. The tilde ($\:\tilde\cdot\:$) over the variables represents the known solutions. It is worth noticing that other genenative models (other than GANs) would also work with the PredGAN algorithm, since it only needs a generator and a way to optimise its output to match known solutions.

Only the time steps from $n-m$ to $n-1$ are taken into account in the functional (Eq.~\eqref{eq:loss_prediction}) which controls the optimisation of $\mathbf{z}^{n}$. After convergence, the newly predicted time level $n$ is added to the known solutions $\tilde{\bm{\alpha}}^{n}$ = $\bm{\alpha}^{n}$, and the converged latent variables $\mathbf{z}^{n}$ are used to initialize the latent variables at the next optimisation to predict time step $n+1$. The process repeats until all time levels are predicted. The gradient of Eq.~\eqref{eq:loss_prediction} is calculated by automatic differentiation \citep{wengert:64, linnainmaa:76, baydin:17}, which means backpropagating the error generated by the loss function in Eq.~\eqref{eq:loss_prediction} through the generator. 

\begin{figure}[htb]
	\centering
	\includegraphics[width=1.0\columnwidth]{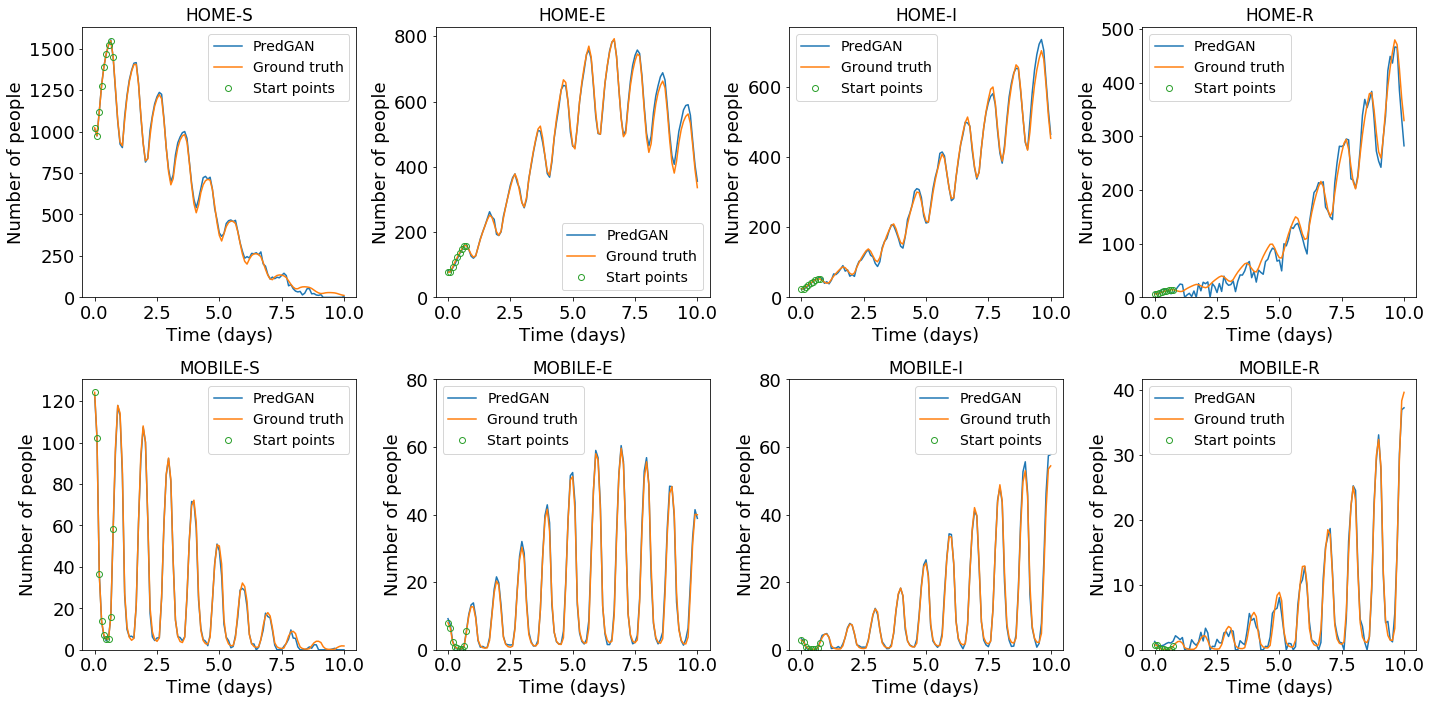}
	\caption{Prediction using the PredGAN. The green circles are the known solutions used to start the prediction process. The blue and orange lines are the prediction and the ground truth, respectively.}
	\label{fig:Predexample}
\end{figure}

Figure \ref{fig:Predexample} shows one example of prediction using the PredGAN. The results represent the evolution, in one cell of the grid (one point in space), of the number of people in each group (home, mobile) and compartment (S, E, I and R) over time. The green circles are the known solutions used to start the prediction process, and the blue and orange lines represent the prediction and the ground truth, respectively. 
Each cycle in the graphs corresponds to a period of one day, when mobile people leave their homes during the day and return at night. After the first nine time iterations the PredGAN does not see any data from the full-order numerical simulation, and relies completely on the predictions from PredGAN. Data from the full-order numerical simulation is only required as a starting point.   

\section{DA-PredGAN for data assimilation}\label{sec:da}

Data assimilation is a type of inverse problem that aims to incorporate observed data into mathematical models. To perform data assimilation using the PredGAN process, \citet{silva:21a} proposed the Data Assimilation Predictive GAN (DA-PredGAN) that involves three changes to the PredGAN. (i) One additional term is included in the loss function in Eq.~\eqref{eq:loss_prediction} to take account of the data mismatch between the observed data and the generated values. (ii) The aim of the data assimilation is to match the observed data and to determine the model parameters $\bm{\mu}^k$ (inputs of the numerical PDE simulation). Therefore, they are not known a priori, as in the prediction. (iii) The forward marching in time is now replaced by forward and backward marching.  

The optimisation at each time step $n$ of the forward march is given by 
\begin{equation}\label{eq:opt_da_forward}
\mathbf{z}^{n}  = \argmin_{\mathbf{z^{n}}} \mathcal{L}_{da}(G(\mathbf{z}^{n})),
\end{equation}
\begin{multline}\label{eq:loss_da_forward}
\mathcal{L}_{da}(G(\mathbf{z}^{n})) =  \sum_{k}^{}\, \left(\tilde{\bm{\alpha}}^k - \bm{\alpha}^k \right)^T \bm{W}_\alpha \left(\tilde{\bm{\alpha}}^k - \bm{\alpha}^k \right)  \\
+  \sum_{k}^{}\,  \zeta_\mu\left(\tilde{\bm{\mu}}^k - \bm{\mu}^k \right)^T \bm{W}_\mu \left(\tilde{\bm{\mu}}^k - \bm{\mu}^k \right) \\
+  \sum_{k}^{}\,  \zeta_{obs}\left(\bm{d}^k - \bm{d}_{obs}^k \right)^T \bm{W}_{obs}^k \left(\bm{d}^k - \bm{d}_{obs}^k \right), 
\end{multline} 
where $k \in \{ n-m, n-m+1, \cdots, n-1\}$. The observed data at each time step $k$ is stored in the vector $\bm{d}_{obs}^k$ of size $N_{obs}$. $\bm{d}^k$ is the generated data calculated based on the output of the generator at time step $k$. In our case, it represents measured data (model states) at some points in the grid and it can be calculated through the POD coefficients $\bm{\alpha}^k$ and stored eigenvectors. $\bm{W}_{obs}^k$ is a square matrix of size $N_{obs}$ whose diagonal values are equal to the observed data weights, and the scalar $\zeta_{obs}$ direct controls how much importance is given to the data mismatch. The values in the diagonal of $\bm{W}_{obs}^k$ are set to zero where we have no observation. After convergence, the newly predicted time level $n$ is added to the known solutions $\tilde{\bm{\alpha}}^{n}$ = $\bm{\alpha}^{n}$, and different from the prediction, we also update the model parameters using the newly predicted time step $\bm{\tilde\mu}^n = \bm{\mu}^n$. 
After the forward march, the process continues with a backward march. For the latter instead of working forward in time, the process goes backwards in time, from the last time step to the first. The loss function for the optimisation at each iteration (backward time step) $n$ of the backward march is defined as in Eq. \eqref{eq:loss_da_forward}, but now for the backward march $k \in \{ n+m, n+m-1, \cdots, n+1\}$. 


After performing a forward and backward march using Eqs.~\eqref{eq:opt_da_forward}, the average of the data mismatch (last term on the right of Eqs.~\eqref{eq:loss_da_forward}) at the end of all iterations $n$ is calculated. If the average mismatch has not converged or the maximum number of iterations is not reached, the process continues with a new forward and backward march. A relaxation factor is also introduced to stabilize the process of marching forward and backward in time as in \citet{silva:21a}.  


\begin{figure}[htb]
	\centering
	\includegraphics[width=1.0\columnwidth]{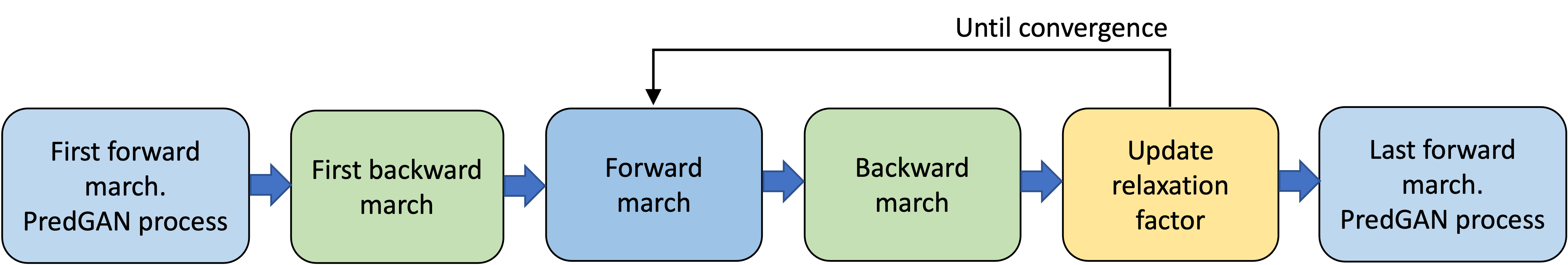}
	\caption{Overview of the DA-PredGAN. Each march represents going through all time steps. The last forward march is optional (mostly for parameterised problems).}
	\label{fig:DA-PredGAN}
\end{figure}

\begin{figure}[!htb]
	\centering
	\begin{subfigure}[t]{1.0\columnwidth}
		\centering
		\includegraphics[width=\columnwidth]{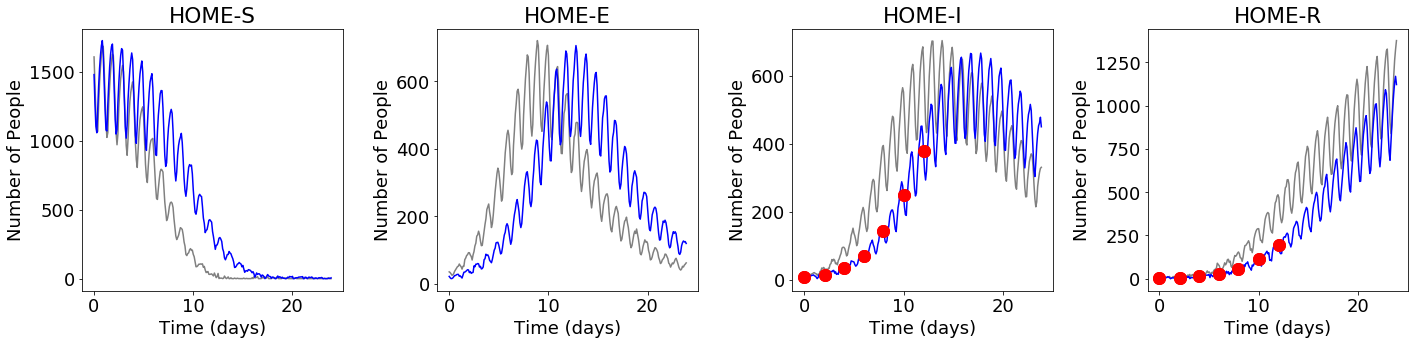}
		\label{fig:prior}
	\end{subfigure}
	\begin{subfigure}[t]{1.0\columnwidth}
		\centering
		\includegraphics[width=\columnwidth]{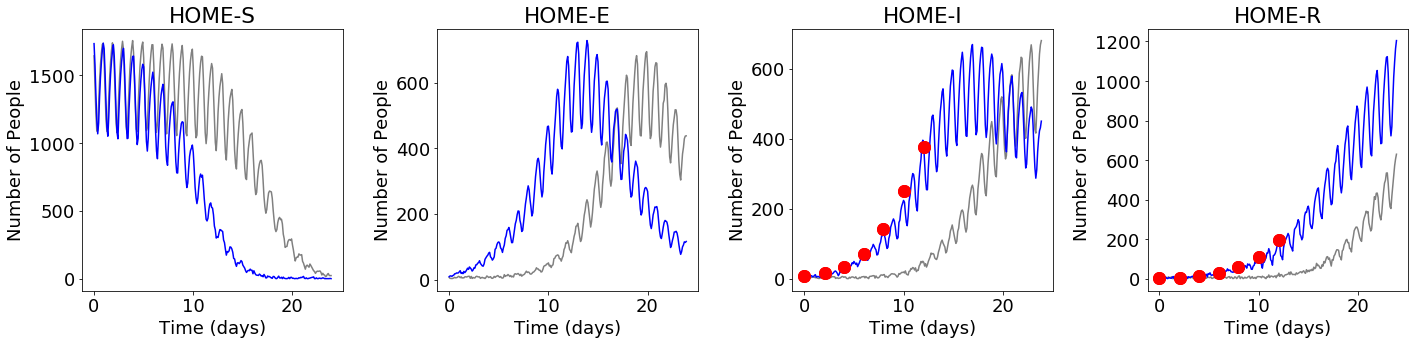}
		\label{fig:post}
	\end{subfigure}
	\caption{Two different data assimilations (top and bottom) using the DA-PredGAN. The circles in red are the observed data (the same in both cases), the lines in grey the initial conditions (first forward march), and the blue lines the final results (last forward march). }
	\label{fig:DA-Predexample}
\end{figure}

Figure \ref{fig:DA-PredGAN} shows an overview of the DA-PredGAN process and Figure \ref{fig:DA-Predexample} shows two data assimilation results (top and bottom plots). The results show the evolution in time of the simulation states in one cell of the grid (one point in space). We generate observed data from a full-order numerical simulation that was not included in the training set of the GAN. 
We also added 5\% noise to the chosen data. We note that for both cases (top and bottom plots) the observed data is honoured, although their results are slightly different. We only show here the home group in order to save space, the results for the mobile group are similar.

\section{UQ-PredGAN for Uncertainty Quantification}\label{sec:uq}

The computation of a single model that matches the observed data is usually insufficient to quantify risks and uncertainties. Data assimilation is generally an ill-posed inverse problem \citep{tarantola:05,oliver:08}, hence several models can match the observed data, within some tolerance (as in Figure \ref{fig:DA-Predexample}). In order to quantify uncertainty, we propose in this work a ROM named Uncertainty Quantification Predictive GAN (UQ-PredGAN). This method is inspired by the RML (or RTO) as a way of sampling a posterior distribution conditioned to observed data. In the RML/RTO, the numerical simulation is used to predict forward, and for each sample, an optimisation (data assimilation) is performed to condition the models to the observed data. The challenge is usually to perform the optimisation, since the full-order numerical simulation needs to be run several times and usually adjoints are not present. In this work, the proposed method UQ-PredGAN can compute uncertainties relying just on a set of unconditioned numerical simulations. The prediction, data assimilation and uncertainty quantification are performed using the inherent adjoint capability present within neural networks, and no additional full-order numerical simulations, other than those used for training the GAN, are required.       

The idea is to generate several models that match the observed data and can quantify the uncertainty in the model states and model parameters. To this end, we perform several data assimilations using the DA-PredGAN algorithm with the modified optimisation process for each forward/backward time step, 
\begin{equation}\label{eq:opt_uq}
\mathbf{z}^{n}  = \argmin_{\mathbf{z^{n}}} \mathcal{L}_{uq,j}(G(\mathbf{z}^{n})),
\end{equation}
\begin{multline}\label{eq:loss_uq}
	\mathcal{L}_{uq,j}(G(\mathbf{z}^{n})) = \sum_{k}^{}\, \left(\tilde{\bm{\alpha}}^k_j - \bm{\alpha}^k \right)^T \bm{W}_\alpha \left(\tilde{\bm{\alpha}}^k_j - \bm{\alpha}^k \right) \\
	+  \sum_{k}^{}\,  \zeta_\mu\left(\tilde{\bm{\mu}}^k_j - \bm{\mu}^k \right)^T \bm{W}_\mu \left(\tilde{\bm{\mu}}^k_j - \bm{\mu}^k \right) \\
	+ \sum_{k}^{}\,  \zeta_{obs}\left(\bm{d}^k - \bm{d}_{obs}^k + \bm{\varepsilon}^k_j\right)^T \bm{W}_{obs}^k \left(\bm{d}^k - \bm{d}_{obs}^k + \bm{\varepsilon}^k_j\right), 
\end{multline}
where for the forward march $k \in \{ n-m, n-m+1, \cdots, n-1\}$ and for the backward march $k \in \{ n+m, n+m-1, \cdots, n+1\}$. Considering $N_s$ as the number of data assimilations to be performed, then $j = 1,...,N_s$. In this work, $N_s=200$. This value was chosen based on previous experience using the RML/RTO. The observed data error is represented by the random vector $\bm{\varepsilon}$, and we consider that all measurement errors are uncorrelated, thus they are sampled from a normal distribution with zero mean and standard deviation equal to $5\%$ of the corresponding observed data. For each data assimilation $j$, we use a different prior $\tilde{\bm{\mu}}^k_j$ with the corresponding initial condition $\{\tilde{\bm{\alpha}}^{k}_{j}\}_{k=0}^{m}$, and a different perturbation on the observed data $\bm{\varepsilon}^k_j$. 

The UQ-PredGAN is proposed as follows: 
\begin{enumerate}
\itemsep0em 
	\item Sample the model parameters $\tilde{\bm{\mu}}_j$ from a normal distribution $\mathcal{N}(\overline{\bm{\mu}},\mathbf{C}_\mu)$, where $\mathbf{C}_\mu$ is the covariance matrix of the model parameters, and $\overline{\bm{\mu}}$ is the model parameter mean vector.  
	\item Sample the measurement error $\bm{\varepsilon}_j$ from a normal distribution $\mathcal{N}(0,\mathbf{C}_d)$, where $\mathbf{C}_d$ is the covariance matrix of the measurement error. 
	\item Assimilate data using the DA-PredGAN process, but using Eqs.~\eqref{eq:opt_uq} and \eqref{eq:loss_uq} for the forward and backward marches. 
	\item Repeat the process until the final number of data assimilation samples $N_s$ is reached. 
\end{enumerate}

After performing $N_s$ steps of the UQ-PredGAN, accept all realizations that obtained an acceptable level of data mismatch. It is worth mentioning that for the RML/RTO, when the case is linear and normal distributions are used to sample the model parameters and measurement error, the RML/RTO samples the corrected posterior distribution \citep{oliver:96,bardsley:14,stordal:18}. In this work, we also use normal distributions to sample the model parameters and measurement error; however, the test case is nonlinear, and the weighting terms are seen as tuning parameters (as in \citet{stordal:18}). Thus, the results are an approximate sample of the posterior distribution.

\section{Extended SEIRS model}\label{sec:appseirs}

The extended SEIRS model used in this work consists of four compartments (Susceptible - Exposed - Infections - Recovered) and two people groups (Home - Mobile). Figure \ref{fig:ExtSEIRSmodel} shows the diagram of how individuals move between compartments and groups. The model starts with some individuals in the infectious compartments (Home-I/Mobile-I). The members of these compartment will  spread the pathogen to the susceptible compartments (Home-S/Mobile-S). Upon being infected, the members of the susceptible compartments are moved to the exposed compartments (Home-E/Mobile-E) and remain there until they become infectious. Infectious individuals remain in the infectious compartment until they become recovered (Home-R/Mobile-R). Recovered people can also become susceptible again due to the loss of immunity.
Modelling the movement of people is of the utmost importance in the spread of infectious diseases, such as COVID-19. Therefore, the goal of the extended SEIRS model is to reproduce the daily cycle of night and day, in which there is a pressure for mobile people to go to their homes at night, and there will be many people leaving their homes during the day and thus joining the mobile group. To this end, the extended SEIRS model uses a diffusion term (last term on the right of Eq.~\ref{eq:Ext-SEIRS}) and an interaction term (penultimate term on the right of Eq.~\ref{eq:Ext-SEIRS}) to model this process:
\begin{subequations}
	\label{eq:Ext-SEIRS} 
	\begin{multline}
	\frac{\partial S_h}{\partial t}  = \eta_h N_h  - \frac{S_h \sum_{h'}(\beta_{h\, h'}   I_{h'}) }{ N_{h}} 
	+  \xi_h R_h - \nu_h^S S_h  \\ - \sum_{h'=1}^{\cal H} \lambda_{h\,h'}^S S_{h'}  
	+ \nabla \cdot ( k_h^S \nabla S_h  ) , 
	\label{eq:Ext-SEIRS-eq1} 
        \end{multline} 
	\begin{multline}
	\frac{\partial E_h}{\partial t}  =  \frac{S_h \sum_{h'}(\beta_{h\, h'}   I_{h'}) }{ N_{h}} 
	-  \sigma_h E_h - \nu_h^E E_h  \\
	- \sum_{h'=1}^{\cal H} \lambda_{h\,h'}^E E_{h'}  + \nabla \cdot ( k_h^E \nabla E_h ),
	\label{eq:Ext-SEIRS-eq2}
        \end{multline}  
	\begin{multline}
	\frac{\partial I_h}{\partial t}   =   \sigma_h E_h - \gamma_h I_h - \nu_h^I I_h 
	- \sum_{h'=1}^{\cal H} \lambda_{h\,h'}^I I_{h'} \\
	+ \nabla \cdot (k_h^I \nabla I_h ),
	\label{eq:Ext-SEIRS-eq3} 
        \end{multline}
	\begin{multline}
	\frac{\partial R_h}{\partial t}   =   \gamma_h I_h  - \xi_h R_h 
	- \nu_h^R R_h
	- \sum_{h'=1}^{\cal H} \lambda_{h\,h'}^R R_{h'}  \\
	+ \nabla \cdot ( k_h^R \nabla R_h ), 
	\label{eq:Ext-SEIRS-eq4} 
	\end{multline}
\end{subequations} 
where ${\cal H}$ represents the number of groups. Here, we have two groups of people, hence ${\cal H}=2$, one representing people at home $h=1$, and the second representing people that are mobile $h=2$ and outside their homes therefore. $N_h$ represents the total number of individuals in each group, $\beta_{h\,h'}$ is the transmission rate between groups, $\sigma_h$ is the rate of exposed individuals becoming infectious, $\gamma_h$ is the recovered rate, and $\xi_h$ is the rate recovered individuals return to the susceptible group due to loss of immunity. The vital dynamics are represented by $\eta_h$ and $\nu_h$, where $\eta_h$ is the birth rate and $\nu_h$ is the death rate. The diffusion coefficient is represented by $k_h$ and describes the movement of people around the domain. The interaction terms, $\lambda_{h\,h'}$, control how people move between groups, for example, how people that are in the mobile group move to the home group. When moving between groups people remain in the same compartment, and when moving between compartments, people remain in the same group. 

One important factor in dynamics of infectious diseases is the basic reproduction number ($\mathcal{R}_0$), which represents the expected number of new cases caused by a single infectious member in a completely susceptible population \citep{dietz:93,hethcote:00}. The $\mathcal{R}_0$ controls how rapidly the disease could spread and for each group it is define as  
\begin{equation}
\label{eq:R0ext}
\mathcal{R}_{0\,h} = \frac{\sigma_h}{(\sigma_h + \nu_h)}\frac{\beta_{h\,h'}}{(\gamma_h+\nu_h)},
\end{equation}
where we assume $\beta_{h\,h'}=0$ when $h \not= h'$ because people in their homes never directly meet mobile people (who are outside their homes). For this case, we can also calculate an Effective $\mathcal{R}_{0}$ representing the $\mathcal{R}_{0}$ seen by the whole population at a specific time. It can be calculated as 
\begin{equation}
    \label{eq:R0ext}
    \text{Effective}~\mathcal{R}_{0} = \frac{\sum_h S_h \mathcal{R}_{0\,h}}{\sum_h S_h},
\end{equation}

\section{Dataset and training process}\label{sec:apptrain}

For the training process 40 full-order numerical simulations were performed in order to generate the training dataset. Each simulation consists of two different $\mathcal{R}_{0\,h}$, one for people at home and another for mobile people. The spatial domain of the numerical simulation is a regular grid of $10\times10$ (100 cells). Considering that each type of people (people at home and mobile) has the four quantities of the extended SEIRS model (susceptible,  exposed, infectious and recovered), there will be eight variables for each cell in the grid per time step, which gives a total number of $100 \times 8 = 800$ variables (model states). Proper orthogonal decomposition is performed in the $800$ variables, in order to work with a low dimensional space in the GN-based ROM. The first $15$ principal components were chosen and they capture $>99.99\%$ of the variance held in the time snapshots. Hence the GAN is trained to generate the $15$ POD coefficients ($\bm{\alpha}^n$) and the two $\mathcal{R}_{0\,h}$ ($\bm{\mu}^n$) over a sequence of 10 time steps. We choose this time length because it represents a cycle (one day) in the results. 

The GAN architecture is based on DCGAN \cite{radford:15} and 
all the codes are implemented using Python and TensorFlow (Apache 2.0 license). 
We choose the size of the latent vector $\mathbf{z}$ to be 100. The networks receive/generate the 10 time levels as a two-dimensional array ("an image", Eq. (\ref{eq:phi})) with 10 rows and 17 columns. Each row represents a time level and each column comprises the $15$ POD coefficients and the two $\mathcal{R}_{0\,h}$. We choose this configuration, instead of a linear representation, to exploit the time dependence captured in the two-dimensional array. We also carried out initial tests using a linear representation of the time level outputs and a multi-layer perceptron as a generator and discriminator. However, it generated worse results than the two-dimensional representation. 

\section{Hardware configuration}\label{sec:apphard}

A Linux (Ubuntu 18.04.6 LTS) workstation was used to train the machine learning models and run all the numerical simulations. Table~\ref{tab:hard} shows the hardware configuration. 

\begin{table}[!htb]
  \caption{Hardware configuration.}
  \label{tab:hard}
  \centering
  \resizebox{\columnwidth}{!}{%
    \begin{tabular}{ll}
    \toprule
        Description & Quantity\\
    \midrule
        16GB DDR4 3200 MHz RAM Memory & 16 \\
        Samsung 970 EVO PLUS 2TB SSD/Solid State Drive  & 1 \\
        Seagate IronWolf PRO 4TB SATA HDD/Hard Drive & 3 \\
        Nvidia Quadro RTX 4000 Video Card & 1 \\
        AMD 32 Core 2nd Gen EPYC 7452 CPU/Processor & 2 \\
        AMD EPYC 7000 EATX Gigabit Server Motherboard & 1 \\
    \bottomrule
    \end{tabular}
    }
\label{table-hyperparameters}
\end{table}

\end{document}